\documentclass{article}

\usepackage[final]{neurips_2024}

\usepackage[utf8]{inputenc} 
\usepackage[T1]{fontenc}    
\usepackage{hyperref}       
\usepackage{url}            
\usepackage{booktabs}       
\usepackage{amsfonts}       
\usepackage{nicefrac}       
\usepackage{microtype}      
\usepackage{xcolor}         
\usepackage{amsmath}
\usepackage{enumitem}
\usepackage{graphicx}
\usepackage{subcaption}

\title{Fostering Intrinsic Motivation in Reinforcement Learning with Pretrained Foundation Models}

\author{%
  Alain Andres$^{1}$\thanks{Corresponding author: https://aklein1995.github.io } \\
  \texttt{alain.andres@tecnalia.com} \\
  \And
  Javier Del Ser$^{1,2}$ \\
  \texttt{javier.delser@tecnalia.com} \\
  \AND
  \textnormal{$^1$ TECNALIA, Basque Research and Technology Alliance (BRTA), Donostia-San Sebastian, Spain} \\
  \textnormal{$^2$ University of the Basque Country, 48013 Bilbao, Spain} \\
}

\begin{document}

\maketitle

\begin{abstract}
Exploration remains a significant challenge in reinforcement learning, especially in environments where extrinsic rewards are sparse or non-existent. The recent rise of foundation models, such as CLIP, offers an opportunity to leverage pretrained, semantically rich embeddings that encapsulate broad and reusable knowledge. In this work we explore the potential of these foundation models not just to drive exploration, but also to analyze the critical role of the episodic novelty term in enhancing exploration effectiveness of the agent. We also investigate whether providing the intrinsic module with complete state information -- rather than just partial observations -- can improve exploration,
despite the difficulties in handling small variations within large state spaces.
Our experiments in the MiniGrid domain reveal that intrinsic modules can effectively utilize full state information, significantly increasing sample efficiency while learning an optimal policy. 
Moreover, we show that the embeddings provided by foundation models are sometimes even better than those constructed by the agent during training, further accelerating the learning process, especially when coupled with the episodic novelty term to enhance exploration.
\end{abstract}

\section{Introduction}\label{sec:intro}
Exploration is a fundamental challenge in reinforcement learning (RL), particularly in environments where extrinsic rewards are sparse or absent. In such scenarios, agents often struggle to efficiently explore and learn, frequently failing to discover meaningful strategies that lead to successful task completion. To address these challenging RL tasks, various methods have been proposed, including approaches that maximize entropy \cite{haarnoja_soft_2018}, techniques that increase the uncertainty of the agent's parameters \cite{fortunato_noisy_2018}, and imitation learning to mimic expert behaviors \cite{zheng_imitation_2024}. Albeit effective, these proposals have their own drawbacks, such as the need for careful configuration tuning, or the requirement of externally gathered data.

Inspired by human psychology, intrinsic motivation encourages agents to be curious (i.e., seek novel situations) in the environment, fostering a more robust and adaptive learning process \cite{aubret_survey_2019}. This typically involves generating an intrinsic reward $r^i_t$ that quantifies the novelty or \emph{surprise}, which is then added to the actual extrinsic reward from the environment $r^e_t$. This addition leads to an updated reward signal $r_t = r^e_t + \beta \cdot r^i_t$, wherein $\beta\in\mathbb{R}^+$ balances between exploiting known information in the environment with fostering the agent's curiosity. However, the effectiveness of these methods often depends on the quality of the state representations used to compute the novelty bonus, which are commonly learned during the agent’s training process.

Recently, foundation models like CLIP (Contrastive Language-Image Pretraining,  \cite{radford_learning_2021}) have gained prominence in the literature for their ability to encode rich, semantically meaningful representations of visual and textual data. These models, pretrained on vast datasets, encapsulate a wide range of reusable knowledge that can be transferred across different tasks and domains. By leveraging these pretrained embeddings, agents can generate more informative intrinsic rewards, driving exploration in a more targeted and meaningful way.

This work builds on FoMoRL (Foundation Models for Semantic Novelty in Reinforcement Learning \cite{gupta_foundation_2022}), which demonstrated the effectiveness of using CLIP embeddings to drive exploration. Our research takes a step further by posing and answering with empirical evidence two main research questions:
\begin{itemize}[leftmargin=*]
    \item How effective are foundation models in generating embeddings that enhance exploration in sparse RL? Can their capabilities be extended by providing full state information?
    
    \textcolor{black}{
    Previous studies have highlighted the challenges of computing novelty from full states \cite{campero_learning_2021}, where methods like Random Network Distillation (RND) \cite{burda_exploration_2018} and Counts \cite{bellemare_unifying_2016} struggle to differentiate states within the same episode (yet increase surprise between episodes). While FoMoRL has shown promise with full state information, it remains uncertain whether more sophisticated techniques like RIDE \cite{raileanu_ride_2020} would also perform competitively in this setup.
    }
    
    \item Is the success of these models driven by the inclusion of an episodic novelty term, which penalizes revisiting states within an episode?
    
    \textcolor{black}{The episodic novelty term has recently been shown to be one of the most important components when generating intrinsic rewards \cite{andres_evaluation_2022,wang_revisiting_2023}. In FoMoRL, it is unclear whether such term was in use, and if so, how it was computed from the RGB images collected by the agent. Decoupling the contribution of foundational model embeddings from the episodic novelty term is crucial for understanding their respective roles.}
\end{itemize}

To address these questions, we conduct a series of experiments over the MiniGrid domain, comparing the performance of agents using different configurations of RIDE and the FoMoRL approach. Throughout this investigation, we aim to provide deeper insights into the mechanisms that drive successful exploration in RL, and explore the potential of foundation models to push the boundaries of open-ended learning.

\section{Related Work}
\paragraph{Intrinsic Motivation for enhanced exploration in RL.} In response to the novelty that the agent experiences during its learning process, the reward signal can be augmented to foster explorative behaviors. This is usually tackled via count-based methods that encourage visits to less frequent states \cite{bellemare_unifying_2016,ostrovski_count-based_2017}, or by reducing uncertainty/prediction error in the environment where the agent is deployed \cite{pathak_curiosity-driven_2017,burda_exploration_2018}. These \textit{lifelong} rewards are significantly beneficial when being combined with \textit{episodic} information \cite{raileanu_ride_2020,flet-berliac_adversarially_2021,zhang_noveld_2022}, being sometimes even more interesting to use solely the latter episodic information \cite{andres_evaluation_2022,henaff_study_2023,wang_revisiting_2023}. Thus, deriving methods that capture this episodic information has become of interest within the community working on sparse RL tasks \cite{savinov_episodic_2019,jo_leco_2022,henaff_exploration_2023}.

\paragraph{Foundational Models and RL.} Recently, some approaches have relied on Vision Language Models (VLM) to generate auxiliary rewards by calculating the cosine similarity between the language goal and the input image \cite{mahmoudieh_zero-shot_2022}. However, their effectiveness seems to be related to the model size and the dataset used for learning them \cite{rocamonde_vision-language_2024,baumli_vision-language_2024}. While these rewards can be noisy, they have proven to be useful during pretraining \cite{adeniji_language_2023} and can be refined through contrastive reward alignment techniques \cite{fu_furl_2024}. Similarly, VLM embeddings have been employed to generate intrinsic rewards, driving exploration more effectively in sparse environments \cite{gupta_foundation_2022}. Additionally, LLMs have been leveraged to guide the learning process, providing high-level instructions \cite{du_guiding_2023} and shaping the reward function \cite{chu_accelerating_2023}.

\section{Methodology}
In this section we motivate the research questions stated in the introduction based on the definition of the rewards at the core of RIDE ad FoMoRL.

\vspace{-3mm}
\paragraph{RIDE: Rewarding Impact-Driven Exploration.} RIDE is one the state-of-the-art exploration strategies proposed for environments with sparse rewards. It computes intrinsic rewards based on the impact an agent's actions have on the environment. Specifically, it measures the difference between consecutive states in a learned state representation space, with larger differences indicating more significant exploration. Mathematically:
\begin{equation}
    \label{eq:ride}
    r^{RIDE}_t(s_t,s_{t+1},a_t) = \frac{\|\phi(s_{t+1}) - \phi(s_{t})\|_2}{\sqrt{N_{\text{ep}}(s_{t+1})}},   
\end{equation}
where $\phi(\cdot)$ refers to the embedding network that is learned during training, $||\cdot||_2$ stands for $L_2$-norm, and $N_{ep}(s_{t+1})$ is the number of times the next state $s_{t+1}$ has been visited.

\vspace{-3mm}
\paragraph{FoMoRL: Foundation Models for Semantic Embeddings.}
FoMoRL proposes a different approach by using pretrained state representations from a foundation model, instead of relying on domain-specific state representations that need to be learned, thereby accelerating exploration.
The intrinsic reward in FoMoRL is computed as:
\begin{equation}
    \label{eq:fomorl}
    r^{FoMoRL}_t(s_t,s_{t+1},a_t) = \frac{\|\text{clip}(s_{t+1}) - \text{clip}(s_{t})\|_2}{\sqrt{N_{\text{ep}}(s_{t+1})}},   
\end{equation}
where $\phi(\cdot)$ is replaced by the outputs of CLIP's visual embeddings.

\vspace{-3mm}
\paragraph{Motivation.} Although Equations \eqref{eq:ride} and \eqref{eq:fomorl} were originally proposed considering the full information about the environment, in practice they have been applied using either the state $s_t$ or the partial observation $o_t$. Furthermore, depending on the environment, this information can be represented as an encoded/compact representation, $\{s_t^{\text{enc}}, o_t^{\text{enc}}\}$, or as an RGB visualization when images are used, \smash{$\{s_t^{\text{rgb}},o_t^{\text{rgb}}\}$}. Our work focuses on understanding the impact of these different representations on exploration. Specifically, we aim to analyze: (1) the effect of using either state or observation in the intrinsic curiosity module, (2) the differences between employing an encoded representation or an image, and (3) the importance of the episodic novelty term in these diverse setups.

\section{Experimental Setup}

\paragraph{MiniGrid.} Experiments are conducted over the MiniGrid benchmark, a collection of procedurally generated grid-based environments that challenge RL agents to explore and navigate effectively in the presence of reward sparsity. 
In MiniGrid, the agent can receive environment information in one of two ways: as a full state, where the agent has complete access to the entire grid layout, $s_t$, or as a partial observation, $o_t$, where the agent’s view is restricted to a limited, egocentric field centered on its current position. Partial observations are thus a subset of the full state, unable to capture the areas behind walls or closed doors. The information, whether full state or partial observation, can be represented in two forms: as a compact encoded vector (e.g., ${s_t^{\text{enc}}, o_t^{\text{enc}}}$) that summarizes the environment's key elements, or as an RGB image (e.g., ${s_t^{\text{rgb}}, o_t^{\text{rgb}}}$). 

\begin{figure}[h]
    \centering
    \includegraphics[width=\linewidth]{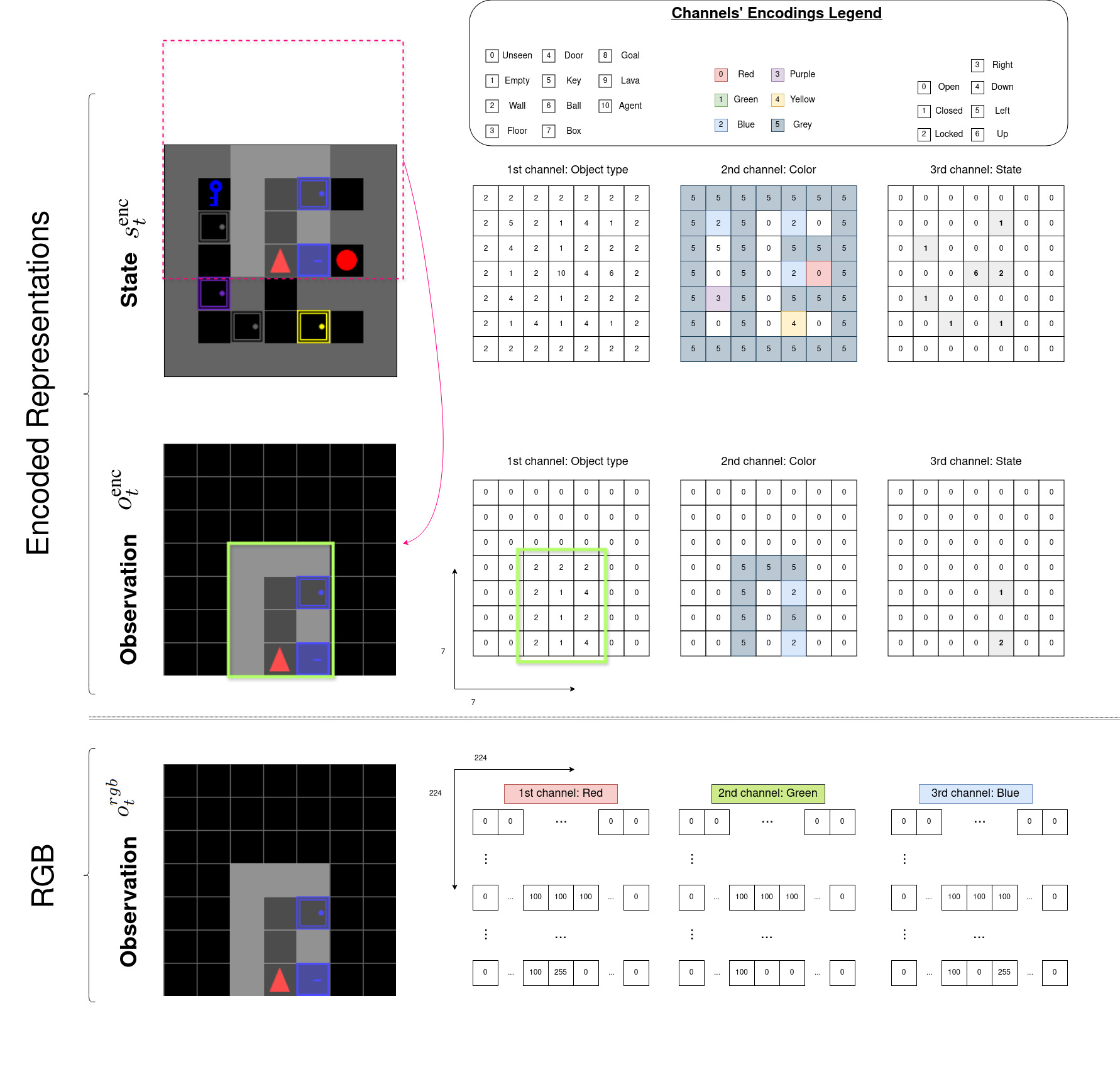}
    \caption{Illustration of full state and partial observation formats in the MiniGrid environment, with two types of representations: encoded and RGB. The top row shows the full state $s_t^{enc}$, where the agent has access to the entire grid layout, represented as a three-channel encoded matrix. The three channels encode object type, color, and state (e.g., open or closed doors, agent orientation). The dotted pink square in this row indicates the region that would compose the agent’s observation in a partial view. The middle row displays this partial observation $o_t^{enc}$, where the agent’s perception is limited to a $7\times7$ egocentric field centered on its position. The green square in this row highlights the area visible to the agent, showing how walls and other objects limit the agent's field of vision. Lastly, the bottom row shows the RGB representation of a partial observation \smash{$o_t^{rgb}$}, where each cell is represented according to three color channels (red, green, and blue) for a pixel-based view.}
    \label{fig:state_observation_compact_rgb}
\end{figure}

Figure \ref{fig:state_observation_compact_rgb} illustrates these different observation types, showing examples of both encoded and RGB representations, as well as full and partial observations. We evaluate performance across several MiniGrid environments, including MultiRoom (\texttt{MN7S4, MN7S8, MN12S10}), KeyCorridor (\texttt{KS3R3, KS4R3}), and ObstructedMaze (\texttt{O2Dlh}).

\vspace{-3mm}
\paragraph{Algorithms.} For training an agent, we use Proximal Policy Optimization (PPO) \cite{schulman_proximal_2017} across all experiments, utilizing RIDE and FoMoRL to generate intrinsic rewards. In our setup, RIDE operates exclusively with encoded information, using either the encoded state $s_t^{enc}$ or observation $o_t^{enc}$ to compute the intrinsic rewards. Conversely FoMoRL, which is pretrained on image data, utilizes RGB representations $s_t^{rgb}$ or $o_t^{rgb}$ for the calculation of its reward. Regardless of the intrinsic motivation technique, when incorporating the episodic novelty term, we use the encoded observation $N_{ep}(o_t^{enc}$) to track state visitation counts. Similarly, the agent's policy is always fed with encoded observations $\pi(\cdot|o_t^{enc}$).
More information about the selected model architectures and hyperparameters can be found in Appendix \ref{app:hyperparamters}.
\begin{table}[t]
    \caption{Performance of RIDE and FoMoRL in terms of the number of required steps to achieve convergence in various MiniGrid environments after training on [\textit{x}M] steps. The (-) indicates that the learning was unsuccessful. The first two columns (Partial) consider only observations for RIDE (\smash{$o_t^{enc}$}) and FoMoRL (\smash{$o_t^{rgb}$}), while the last two columns (Full) consider state information for RIDE ($s_t^{enc}$) and FoMoRL (\smash{$s_t^{rgb}$}). All the results make use of the episodic term component, \smash{$\sqrt{N_{ep}(o_t^{enc})}$}.}
    \vspace{2mm}
    \centering
    \label{tab:results_obs_state}
    \begin{tabular}{lccccc}
    \toprule
    \textbf{} & \multicolumn{2}{c}{Partial, $o_t$} & & \multicolumn{2}{c}{Full, $s_t$} \\ \cmidrule{2-3} \cmidrule{5-6}
    \textbf{} & RIDE & FoMoRL & & RIDE & FoMoRL \\ \midrule
    \texttt{MultiRoom-N7-S4} [10M]       & \textbf{1.53M} & 4.03M & & 0.93M & \textbf{0.58M} \\
    \texttt{MultiRoom-N7-S8} [20M]       & - & - & & - & \textbf{2.8M} \\
    \texttt{MultiRoom-N12-S10} [30M]     & - & - & & - & \textbf{4.06M} \\
    \texttt{KeyCorridorS3R3} [30M]       & 6.82M & \textbf{5.29M} & & \textbf{2.07M} & 5.13M \\ 
    \texttt{KeyCorridorS4R3} [50M]       & - & \textbf{17.06M} & & \textbf{7.92M} & 8.15M \\ 
    \texttt{ObstructedMaze-2Dlh} [50M]   & 39.03M & \textbf{19.92M} & & 29.02M & \textbf{16.53M} \\ 
    \bottomrule
    \end{tabular}
    \vspace{-3mm}
\end{table}

\section{Results}

\paragraph{Impact of Partial vs Full Observability.}
Table \ref{tab:results_obs_state} provides the number of steps required for RIDE and FoMoRL to converge to the optimal policy in different MiniGrid environments, considering both partial observations and full state information. For both approaches, \emph{the use of full state information significantly augments the sample-efficiency}.
In complex \texttt{MultiRoom} environments only FoMoRL with full state access manages to effectively train. However, in \texttt{KeyCorridor},
RIDE performs slightly better than FoMoRL. We hypothesize that this is because these environments contain a higher diversity of objects, allowing RIDE to effectively distinguish between different states and learn a better representation. In contrast, FoMoRL relies on pretrained embeddings from CLIP, which remain frozen during training and may lack the adaptability required to fully capture the variations in these more diverse environments.

\paragraph{Role of the Episodic Term.} Table \ref{tab:results_noep} highlights the critical importance of the episodic term in accelerating the learning process. Without this term, the sample efficiency is significantly reduced. In fact, the absence of the episodic term has a particularly negative impact on RIDE’s performance, as it is unable to learn in \texttt{MultiRoom-N7-S8} and \texttt{MultiRoom-N12-S10}. FoMoRL, however, demonstrates greater resilience to the removal of the episodic term, likely due to its reliance on latent knowledge from pretrained embeddings. Nonetheless, akin to the results with the episodic term, FoMoRL still obtains slightly worse results in some environments.
\begin{table}[h]
    \caption{Performance of RIDE and FoMoRL in terms of the number of required steps to achieve convergence, without considering the episodic term. That is, $\forall o_t$ $\xrightarrow{}$ $\sqrt{N_{ep}(o_t)} = 1$ in Expressions \eqref{eq:ride} and \eqref{eq:fomorl}.}
    \vspace{2mm}
    \centering
    \label{tab:results_noep}
    \begin{tabular}{lcc}
    \toprule
    \textbf{} & RIDE (Full, $s_t^{enc}$) & FoMoRL (Full, $s_t^{rgb}$) \\ \midrule
    \texttt{MultiRoom-N7-S4} [20M]        & \textbf{3.19M} & 14.92M \\
    \texttt{MultiRoom-N7-S8} [30M]        & -              & \textbf{11.89M} \\
    \texttt{MultiRoom-N12-S10} [50M]      & -              & \textbf{25.32M} \\
    \texttt{KeyCorridorS3R3} [50M]        & \textbf{5.27M} & 7.12M \\
    \texttt{KeyCorridorS4R3} [50M]        & \textbf{11.95M} & - \\
    \texttt{ObstructedMaze-2Dlh} [80M]    & 24.76M & \textbf{15.54M} \\ 
    \bottomrule
    \end{tabular}
\end{table}

We refer to Appendix \ref{app:graphic_results} for more details about the results summarized in Tables \ref{tab:results_obs_state} and \ref{tab:results_noep}.

\section{Conclusions}
We have investigated the use of foundation models and episodic novelty to enhance exploration in sparse RL environments. 
Our results evince that providing full state information to the intrinsic module significantly accelerates the agent's learning process, enabling faster convergence to learn optimal policies. 
The episodic novelty term is critical for effective exploration; without it, agents in some environments struggle to learn. However, when prelearned state embeddings are used (as in FoMoRL), access to full state information  can reduce the dependence on the episodic term. 
This benefit is lessened if the embedding must be learned during training (as in RIDE), as the initial exploration can hinder the characterization of suitable state representations.

In the future we plan to extend our approach to more MiniGrid environments, other RL benchmarks with visual inputs, and explore alternatives such as alignment approaches that build on top of the embeddings provided by CLIP and other foundation models.

\newpage
\bibliographystyle{plainnat}
\bibliography{bibliography}

\newpage
\appendix

\section{Model Architectures and Hyperparameters} \label{app:hyperparamters}
\paragraph{Intrinsic motivation.} We conducted a grid search over the range $\beta\in\{0.1, 0.05, 0.01, 0.005, 0.001, 0.0005, 0.0001\}$ to determine the best scaling factor for $r_t^i$ in each environment. Table \ref{tab:gs_partial_full_episodic} summarizes the selected coefficients when considering both full and partial information with the inclusion of the episodic term. Table \ref{tab:gs_noepisodic} provides the chosen values when the episodic component is not used.
\begin{table}[h]
    \caption{\(\beta\) coefficient with episodic counts.}
    \vspace{2mm}
    \centering
    \label{tab:gs_partial_full_episodic}
    \begin{tabular}{lccccc}
    \toprule
    \textbf{} & \multicolumn{2}{c}{Partial} & & \multicolumn{2}{c}{Full} \\ \cmidrule{2-3} \cmidrule{5-6}
    \textbf{} & RIDE & FoMoRL & &  RIDE & FoMoRL \\ \midrule
    \texttt{MultiRoom-N7-S4}        & 0.005 & 0.0005 & & 0.05 & 0.005 \\
    \texttt{MultiRoom-N7-S8}        & 0.05 & 0.005 & & 0.05 & 0.005 \\
    \texttt{MultiRoom-N12-S10}      & 0.01 & 0.01 & & 0.01 & 0.01 \\
    \texttt{KeyCorridorS3R3}        & 0.005 & 0.0005 & & 0.05 & 0.005          \\ 
    \texttt{KeyCorridorS4R3}        & 0.1 &  0.0005 & &  0.1 &  0.001          \\ 
    \texttt{ObstructedMaze-2Dlh}    & 0.005 & 0.0005 & & 0.005 & 0.0005          \\ 
    \bottomrule
    \end{tabular}
\end{table}

\begin{table}[h]
    \caption{\(\beta\) coefficient without episodic counts, i.e., $\forall o_t$ $\xrightarrow{}$ $\sqrt{N_{ep}(o_t)} = 1$ in Equations \eqref{eq:ride} and \eqref{eq:fomorl}.}
    \vspace{2mm}
    \centering
    \label{tab:gs_noepisodic}
    \begin{tabular}{lcc}
    \toprule
    \textbf{} & \multicolumn{2}{c}{Full} \\ 
    \cmidrule{2-3}
    \textbf{} & RIDE & FoMoRL \\
    \midrule
    \texttt{MultiRoom-N7-S4}        & 0.05 & 0.005 \\
    \texttt{MultiRoom-N7-S8}        & 0.05 & 0.005 \\
    \texttt{MultiRoom-N12-S10}      & 0.01 & 0.01 \\
    \texttt{KeyCorridorS3R3}        & 0.05 & 0.0005 \\ 
    \texttt{KeyCorridorS4R3}        & 0.1  & 0.0005 \\ 
    \texttt{ObstructedMaze-2Dlh}    & 0.005 & 0.0005\\ 
    \bottomrule
    \end{tabular}
\end{table}

\paragraph{PPO.} We used a discount factor $\gamma = 0.99$, a clipping factor $\epsilon = 0.2$, 4 epochs per training step, and $\lambda = 0.95$ for GAE. We employed 16 parallel environments to collect rolloutsS of size 128, resulting in a total horizon of 2,048 steps between updates.

\paragraph{Neural Network Architecture.} We used an actor-critic architecture with shared weights between the actor and critic networks. The shared network consisted of three convolutional layers (each with 32 filters of size $3\times 3$, a stride of 2, and padding of 1), followed by a fully connected layer with 256 units. This shared representation was then fed into two separate heads: one for the actor (policy) and another for the critic (value function).

\newpage
\section{Graphical Results} \label{app:graphic_results}

\begin{figure}[h]
    \centering
    \includegraphics[width=\linewidth]{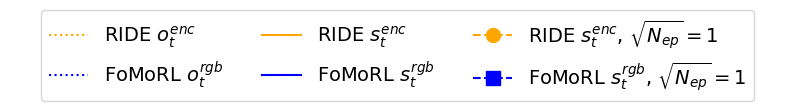}
    \subfloat[MultiRoom-N7-S4]{\includegraphics[width=0.45\linewidth]{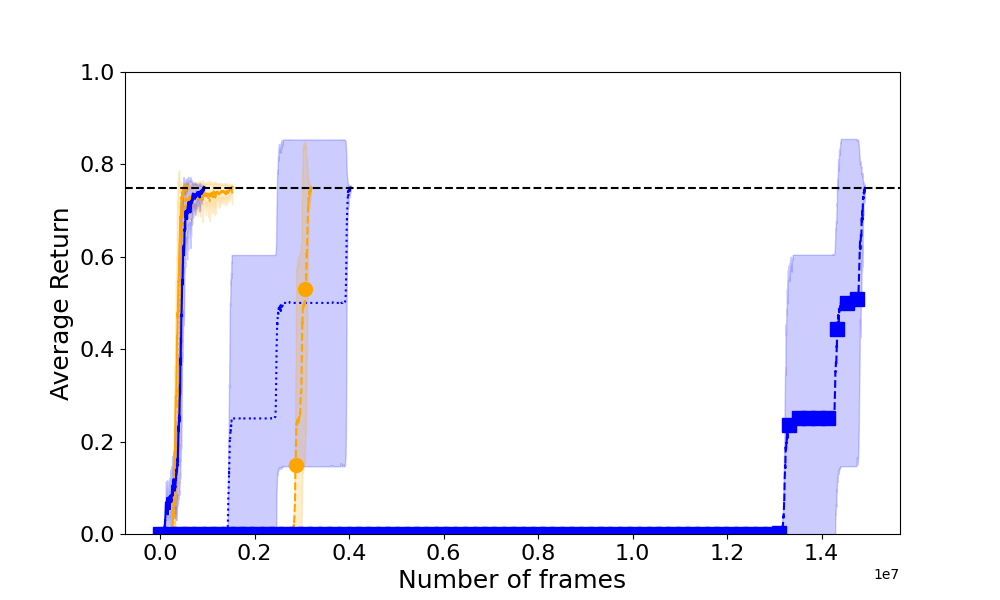}}
    \subfloat[MultiRoom-N7-S8]{\includegraphics[width=0.45\linewidth]{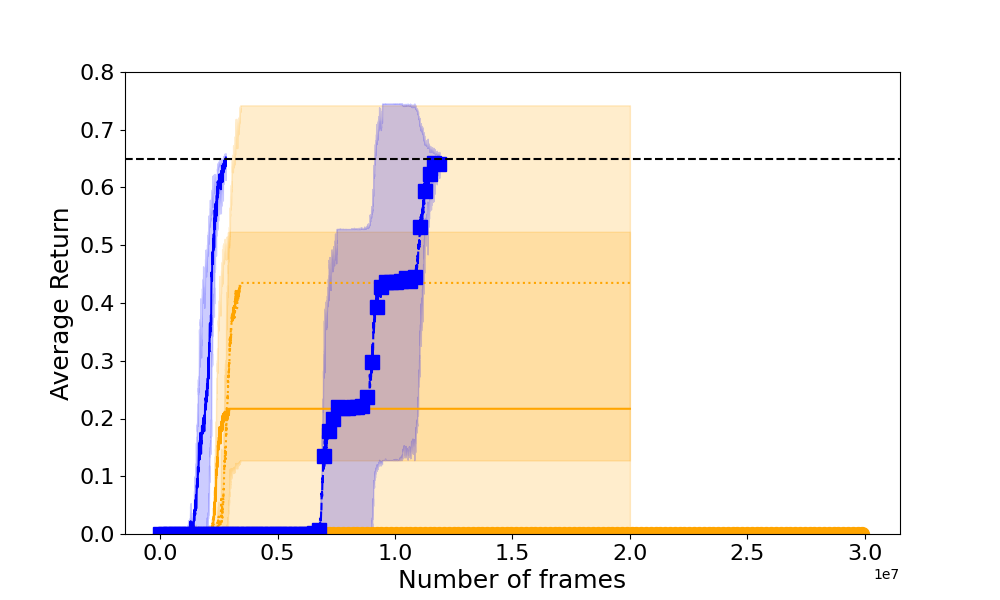}}

    \subfloat[MultiRoom-N12-S10]{\includegraphics[width=0.45\linewidth]{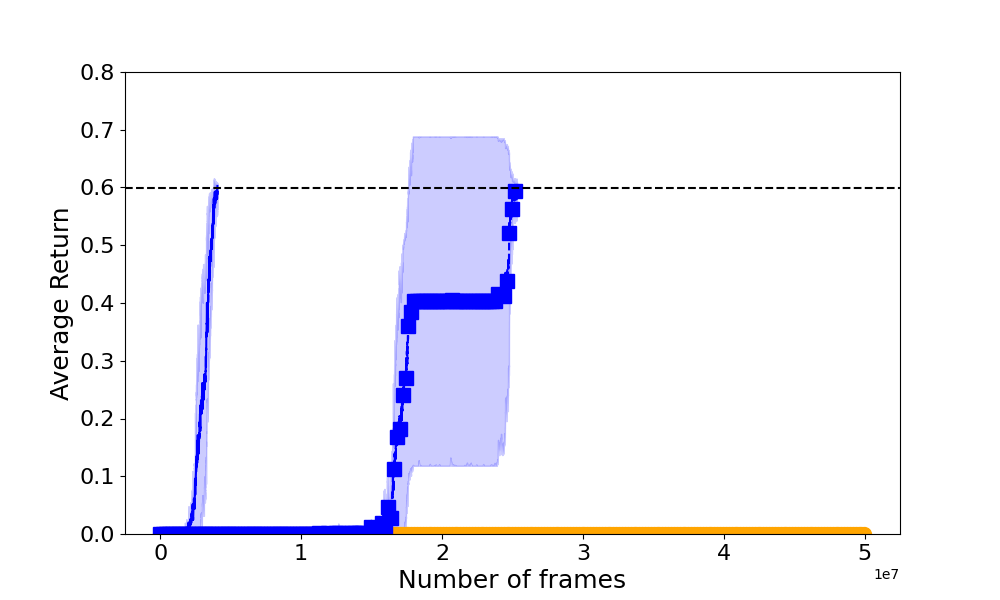}}
    \subfloat[ObstructedMaze-2Dlh]{\includegraphics[width=0.45\linewidth]{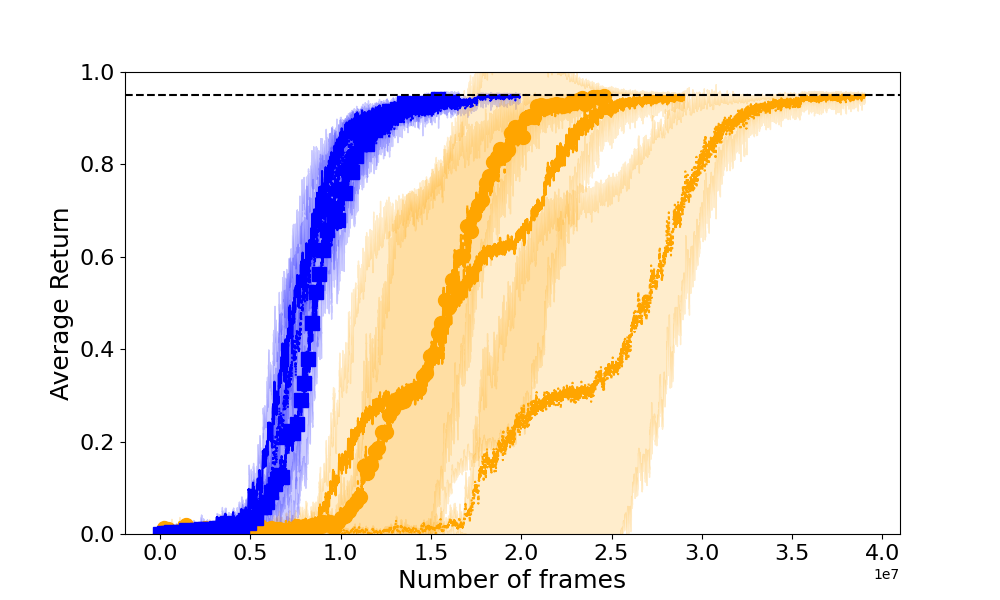}}
    
    \subfloat[KeyCorridorS3R3]{\includegraphics[width=0.45\linewidth]{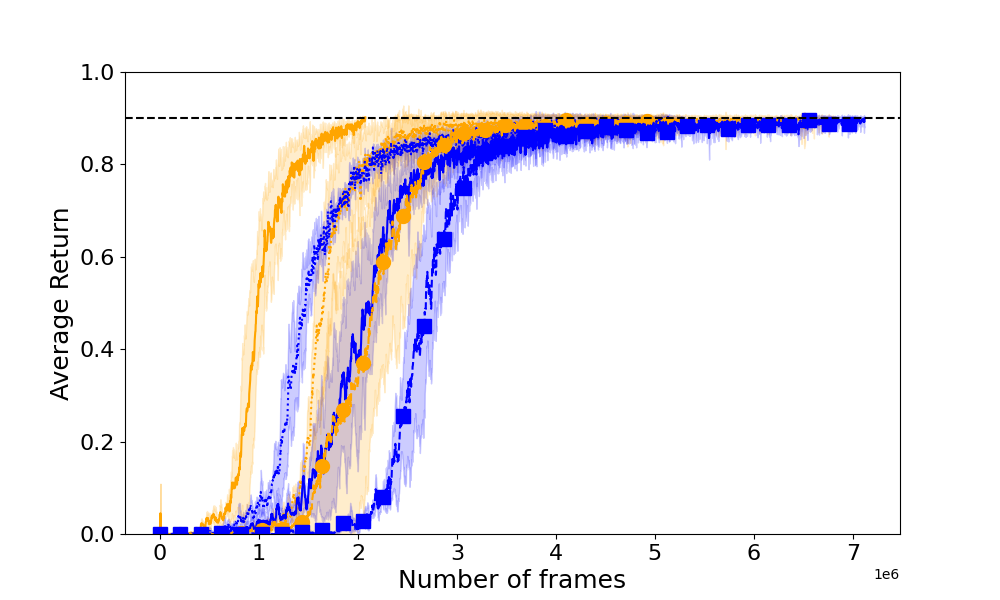}}
    \subfloat[KeyCorridorS4R3]{\includegraphics[width=0.45\linewidth]{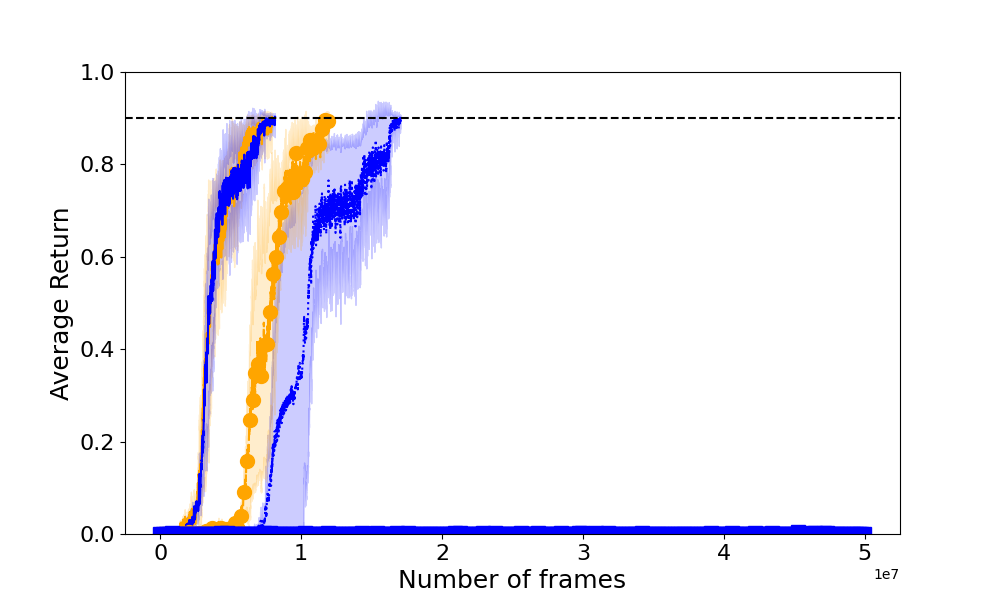}}

    \caption{Performance comparison between RIDE (\textcolor{orange}{orange}) and FoMoRL (\textcolor{blue}{blue}) across various MiniGrid environments and input types. Each subfigure presents the average return over training steps, illustrating learning progress and convergence speed for each algorithm in specific settings. Solid lines represent full observations (\(s_t\)), while dotted lines denote partial observations (\(o_t\)). The horizontal black dashed line indicates the expected return of the optimal policy. The shaded area represents the standard deviation computed across 3 different seeds.}
    \label{fig:enter-label}
\end{figure}

\end{document}